\documentclass[final]{cvpr}

\usepackage{times}
\usepackage{epsfig}
\usepackage{graphicx}
\usepackage{amsmath}
\usepackage{amssymb}
\usepackage{bm}
\usepackage{multirow}
\usepackage{soul}
\usepackage{subcaption}

\newcommand{\Tref}[1]{Table~\ref{#1}}

\newcommand{\Fref}[1]{Figure~\ref{#1}}

\newlength\savewidth\newcommand\shline{\noalign{\global\savewidth\arrayrulewidth
  \global\arrayrulewidth 1pt}\hline\noalign{\global\arrayrulewidth\savewidth}}

\usepackage[pagebackref=true,breaklinks=true,colorlinks,bookmarks=false]{hyperref}

\def\confYear{CVPR 2021}

\begin{document}

\title{Unsupervised Pre-training for Person Re-identification}

\author{Dengpan Fu$^1$ \quad
		Dongdong Chen$^2$ \quad
		Jianmin Bao$^2$\thanks{Corresponding author.} \quad
		Hao Yang$^2$ \\
		Lu Yuan$^2$ \quad
		Lei Zhang$^2$ \quad
		Houqiang Li$^1$ \quad
		Dong Chen$^2$
		\and 
		$^1$University of Science and Technology of China \quad $^2$Microsoft Research  \\ 
		{\tt\small fdpan@mail.ustc.edu.cn} \quad {\tt\small cddlyf@gmail.com} \quad {\tt\small lihq@ustc.edu.cn} \\
		{\tt\small \{jianbao,haya,luyuan,leizhang,doch\}@microsoft.com}
	}

\maketitle 
\begin{abstract}
In this paper, we present a large scale unlabeled person re-identification (Re-ID) dataset ``LUPerson" and make the first attempt of performing unsupervised pre-training for improving the generalization ability of the learned person Re-ID feature representation. This is to address the problem that all existing person Re-ID datasets are all of limited scale due to the costly effort required for data annotation. Previous research tries to leverage models pre-trained on ImageNet to mitigate the shortage of person Re-ID data but suffers from the large domain gap between ImageNet and person Re-ID data. LUPerson is an unlabeled dataset of 4M images of over 200K identities, which is $30\times$ larger than the largest existing Re-ID dataset. It also covers a much diverse range of capturing environments (\eg, camera settings, scenes, etc.). Based on this dataset, we systematically study the key factors for learning Re-ID features from two perspectives: data augmentation and contrastive loss. Unsupervised pre-training performed on this large-scale dataset effectively leads to a generic Re-ID feature that can benefit all existing person Re-ID methods. Using our pre-trained model in some basic frameworks, our methods achieve state-of-the-art results without bells and whistles on four widely used Re-ID datasets: CUHK03, Market1501, DukeMTMC, and MSMT17. Our results also show that the performance improvement is more significant on small-scale target datasets or under few-shot setting. Code is available at: \url{https://github.com/DengpanFu/LUPerson}.
\end{abstract}

\section{Introduction}
\label{sec:intro}

\begin{figure}[t]
\begin{center}
    \subfloat[Market1501]{\includegraphics[width=1.62in]{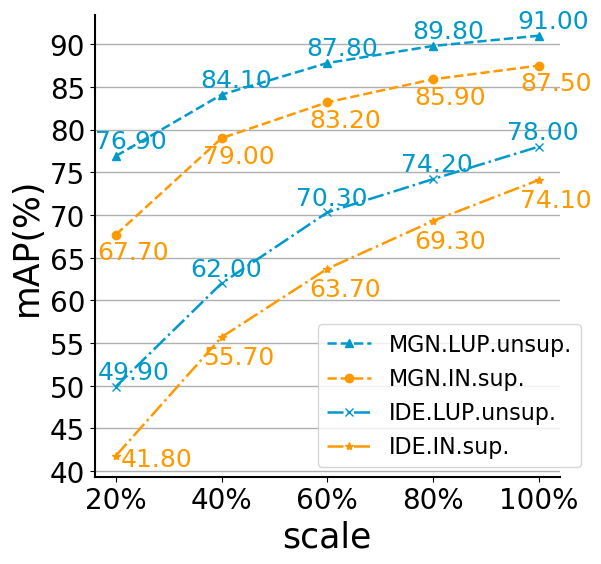}}
    \subfloat[DukeMTMC]{\includegraphics[width=1.62in]{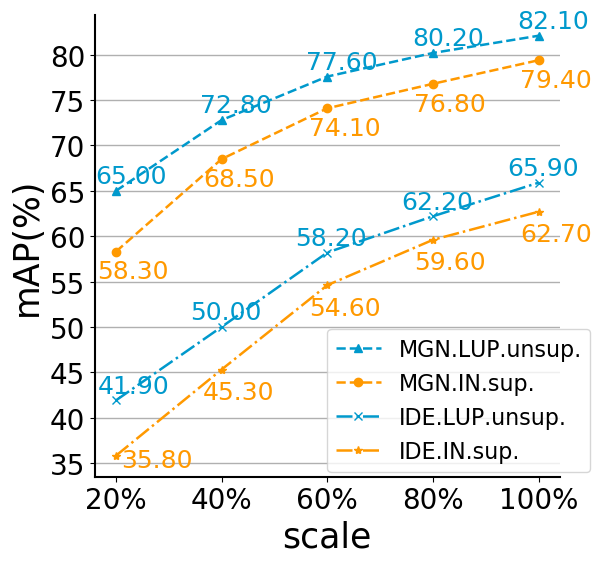}}
\end{center}
\vspace{-0.4cm}
\caption{Person Re-ID performance comparison of applying different pre-trained models on two methods: IDE~\cite{zheng2017person} and MGN~\cite{wang2018learning}. We report the results on different dataset scales for Market1501 and DukeMTC with small-scale setting. IN sup. and LUP unsup. are the supervised model trained on ImageNet and the unsupervised model trained on LUPerson, respectively.}
\vspace{-0.5cm}
\label{fig:onecol}
\end{figure}

Model pre-training plays an indispensable role in person Re-identification (Re-ID). Compared to other vision tasks, as data collection and annotation for Re-ID is extremely difficult and expensive, existing public datasets all have a limited scale in terms of image number (largest MSMT17~\cite{wei2018person}, 126K images), person identities (largest Airport~\cite{karanam2016comprehensive}, 9,651 identities) and captured environments ($<20$ scenes, fixed cameras and resolution). To mitigate the shortage of person Re-ID data, previous research has tried to leverage models pre-trained on ImageNet and transfer the pre-trained feature to Re-ID tasks ~\cite{hermans2017defense, chen2018group, suh2018part, wang2018learning,fu2020improving}. However, it is arguable if using ImageNet for pre-training is optimal, due to the large domain gap between ImageNet and person Re-ID data.

Inspired by the recent success of self-supervised learning ~\cite{chen2020simple,chen2020mocov2,he2020momentum}, we make the first attempt towards large scale unsupervised pre-training for person Re-ID feature representation learning in this paper. Considering the limited scale of existing Re-ID datasets, we build a new \textbf{L}arge-scale \textbf{U}nlabeled \textbf{Person} Re-ID dataset ``\textbf{LUPerson}". It consists of 4M person images of over 200K identities extracted from 46K YouTube videos, which is $\mathbf{30\times}$ larger than the largest existing Re-ID dataset MSMT~\cite{wei2018person}. Moreover, the collected videos cover a wide range of capturing environments (\eg, using fixed or moving cameras, under dynamic scenes, or having different resolutions), yielding a great data diversity which is essential for learning generic representation.
We hope this study and the developed LUPerson dataset will serve as a solid baseline and motivate more feature representation learning researches for person re-identification.

Based on the LUPerson dataset, we systematically study the problem of unsupervised Re-ID feature learning. We find that directly applying the commonly used contrastive learning method such as MoCo v2 \cite{chen2020improved} does not work well for the Re-ID task. After a careful investigation, we discover some  unique factors when applying unsupervised pre-training to the Re-ID task: 1) As a common data augmentation~\cite{he2020momentum,chen2020big,caron2020unsupervised}, the color distortion (\eg, color jitter) is harmful to Re-ID feature learning. This is because the color information is a crucial clue for Re-ID. 2) To prevent contrastive learning from degrading to a trivial solution,  a strong task-specific augmentation operation RandomErasing~\cite{zhong2020random} is proved as beneficial as in supervised Re-ID training. 3) How to use a proper temperature parameter in the contrastive loss plays an important role in finding a balance between  maintaining the discriminativity and mining the hard negatives.

We demonstrate the effectiveness of our pre-trained model on various person Re-ID datasets. Upon the strong MGN~\cite{wang2018learning} baseline, our pre-trained model can improve the \emph{mAP} by $3.5\%$ on Market1501~\cite{Zheng2015ScalablePR}, $2.7\%$ on DukeMTMC~\cite{zheng2017unlabeled}, and $2.0\%$ on MSMT17\cite{wei2018person}; while achieving $2.9\%$ \emph{mAP} gain on CUHK03~\cite{Zheng2015ScalablePR} based on another strong BDB~\cite{dai2019batch} baseline. These results are superior to all the state-of-the-art methods. Our results show that the performance improvement is even more significant with small-scale training samples for different datasets and baselines, as shown in Fig.\ref{fig:onecol}. Besides, our pre-trained model is also general to unsupervised Re-ID methods. Based on the strongest baseline SpCL~\cite{ge2020selfpaced}, our pre-trained model consistently achieves remarkable improvements on various datasets. To the best of our knowledge, this is the first showing that large scale unsupervised pre-training can significantly benefit the person Re-ID task.

Our key contributions can be summarized as follows:
\begin{itemize}
    \item We build a large-scale unlabeled dataset LUPerson, which consists of 4M images for over 200K identities, for unsupervised person Re-ID feature learning. This dataset is much larger than any existing public datasets and enables the first unsupervised pre-training for person Re-ID tasks.
    \item We make generic unsupervised pre-training possible for Re-ID tasks, by carefully investigating the crucial factors, such as data augmentation strategies and the temperature usage in the contrastive learning framework.
    \item The unsupervised representation learning is general to not only supervised Re-ID methods, but also un-supervised Re-ID methods, and helps significantly improve their performance on different datasets.
\end{itemize}

\section{Related Work}
\label{sec:related}

\noindent\textbf{Supervised Person Re-ID.} 
Most existing Re-ID approaches are based on supervised learning on labeled datasets. There are three typical categories of approaches: 1) learning a global feature from the whole image, with supervision imposed through a classification loss, \eg, IDE model~\cite{zheng2017person} and ~\cite{shen2018person}; 2) using a hard triplet loss on the global feature to ensure a smaller distance for features of the same person, such as \cite{hermans2017defense}; 3) learning a part-based feature instead. For example, Sun \etal \cite{sun2018PCB} proposed to partition an image feature into multiple horizontal strips each learned with a separate classification loss, and Suh \etal \cite{suh2018part} presented a part-aligned bi-linear representations. The MGN \cite{wang2018learning}, known as one of the state-of-the-art (SOTA) methods, combined both a classification loss on the global feature and a triplet loss on the local features. Our pre-trained model can be used in the above three representative methods, and show better performance and generalization ability. 

\noindent\textbf{Unsupervised Person Re-ID.} 
Recently, some works attempted to directly train an unsupervised Person Re-ID model without utilizing any labels on existing Re-ID datasets. BUC \cite{lin2019bottom} proposes a bottom-up hierarchical clustering method to jointly optimize a network and the relationship among samples. Mutual Mean-Teaching (MMT)~\cite{ge2019mutual} adopts two networks and learns mutually to refine the hard and soft pseudo labels in the target domain to mitigate the effects of noisy pseudo labels. MMCL \cite{wang2020unsupervised} formulates unsupervised person Re-ID as a multi-label classification task to progressively seek true labels. SpCL \cite{ge2020selfpaced} is known as the SOTA approach in this literature. Its key idea is to fine-tune the network using pseudo labels generated from reliable clustering results trained with contrastive losses. In contrast to these methods, our work focuses on the unsupervised pre-training phase to learn a feature representation that can be generalized to either supervised or unsupervised Re-ID approaches. 

\noindent\textbf{Unsupervised Representation Learning.} Benefit from contrastive learning~\cite{wu2018unsupervised, chen2020simple, he2020momentum}, unsupervised pre-training can learn feature representations with comparable quality to that learned from supervised approaches. Specifically, Wu \etal \cite{wu2018unsupervised} proposed to store representations within a memory bank. MoCo and MoCo v2 \cite{he2020momentum,chen2020improved} introduced a dynamic queue to maintain slowly updated representations to generate negative samples. On the contrary, SimCLR and SimCLR v2 \cite{chen2020simple,chen2020big} proved that a projection head and rich data augmentations can also lead to advantageous visual representations even without these memory structures. 
Recently, BYOL \cite{grill2020bootstrap} shows a good performance even without using any negative pairs. In this paper, we choose the MoCo v2 \cite{chen2020mocov2} framework as our unsupervised pre-training method. Unfortunately, directly applying this framework on person Re-ID tasks does not work well, which requires an in-depth study to identify the key factors which make Re-ID tasks different from generic visual representation learning.

\begin{table*}[t]
\setlength{\tabcolsep}{1.3mm}
    \centering
    \begin{tabular}{c|c|c|c|c|c|c|c|c}
    \hline
        Datasets & \#images & \#scene & \#persons & environment & camera view & resolution & detector & crop size \ \\
        \hline
        \hline
        VIPeR\cite{gray2008viewpoint} & 1,264 & 2 & 632 & - & fixed & fixed & hand & $128\times48$ \\
        GRID\cite{loy2013person} & 1,275 & 8 & 1,025 & subway & fixed & fixed & hand & vary \\
        CUHK03\cite{li2014deepreid} & $14,096$ & $2$ & $1,467$ & campus & fixed & fixed & DPM\cite{felzenszwalb2009object}+hand & vary\\
        Market\cite{Zheng2015ScalablePR} & $32,668$ & $6$ & $1,501$ & campus & fixed & fixed & DPM\cite{felzenszwalb2009object}+hand & $128\times64$ \\
        Airport\cite{karanam2016comprehensive} & $39,902$ & $6$ & $9,651$ & airport & fixed & fixed & ACF\cite{dollar2014fast} & $128\times64$ \\
        DukeMTMC\cite{zheng2017unlabeled} & $36,411$ & $8$ & $1,852$ & campus & fixed & fixed & Hand &  vary \\
        MSMT17\cite{wei2018person} & $126,441$ & $15$ & $4,101$ & campus & fixed & fixed & FasterRCNN\cite{ren2015faster} & vary \\
        \hline
        LUPerson & 4,180,243 & 46,260 & $>200k$ & vary & dynamic & dynamic & YOLOv5 & vary \\
        \hline
    \end{tabular}
    \vspace{-0.5em}
    \caption{The statistics comparison between existing popular Re-ID datasets and our large scale LUPerson dataset. It shows that LUPerson is the current largest Re-ID dataset and has much better diversity.}
    \label{tab:data-stat}
\end{table*}

\section{LUPerson: Large-scale Re-ID Dataset}
\label{sec:LUPerson}

Data is the life-blood of training deep neural network models and ensuring their success. For the person Re-ID task, sufficient and high-quality data are also indispensable for increasing the model's generalization capability. A good Re-ID dataset requires not only a large amount of identities, each appearing in multiple cameras, but also a great diversity in terms of pose variation and capturing environments (\eg, camera angles, resolutions, scenes, etc.). Unfortunately, developing and annotating such a large-scale Re-ID dataset is extremely difficult and expensive.

All existing Re-ID datasets are of limited scale and diversity. \Tref{tab:data-stat} lists the statistics of existing popular Re-ID datasets, including VIPeR~\cite{gray2008viewpoint}, GRID~\cite{loy2013person}, CUHK03~\cite{li2014deepreid}, Market-1501~\cite{Zheng2015ScalablePR}, Airport~\cite{karanam2016comprehensive}, DukeMTMC~\cite{zheng2017unlabeled}, and MSMT17~\cite{wei2018person}. As we can see, the largest dataset only consists of less than $0.2$M images and the largest number of identities is less than 10K. Moreover, these datasets cover very limited scenes ($<20$) and camera settings. As a result, it is hard to use such datasets to learn a high-quality feature representation which is as good as that learned in the generic image classification task which can utilize 1.2M images (\ie, ImageNet-1k) or even 14M images (\ie, ImageNet-22k).

In this work, we ask the question: \textit{can we develop a person Re-ID dataset which is as large as ImageNet?} To address this question, we build \textbf{LUPerson}: a \textbf{L}arge-scale \textbf{U}nlabeled \textbf{Person} Re-ID dataset, consisting of 4M images for more than 200K identities collected from 46K scenes. To our best knowledge, this is the largest scale person Re-ID dataset. 

The development of LUPerson is inspired by the recent success of unsupervised pre-training on generic image classification tasks. Although LUPerson is an unlabeled dataset, the large number of identities and diverse capturing environments included in the dataset offer a great potential for learning a high-quality person Re-ID feature representation that can be utilized to boost the performance of all kinds of Re-ID CNN models. We will make the dataset publicly available to motivate more research works to advance the state of the art of person re-identification.

\begin{figure}[t]
\begin{center}
    \includegraphics[width=1.\linewidth]{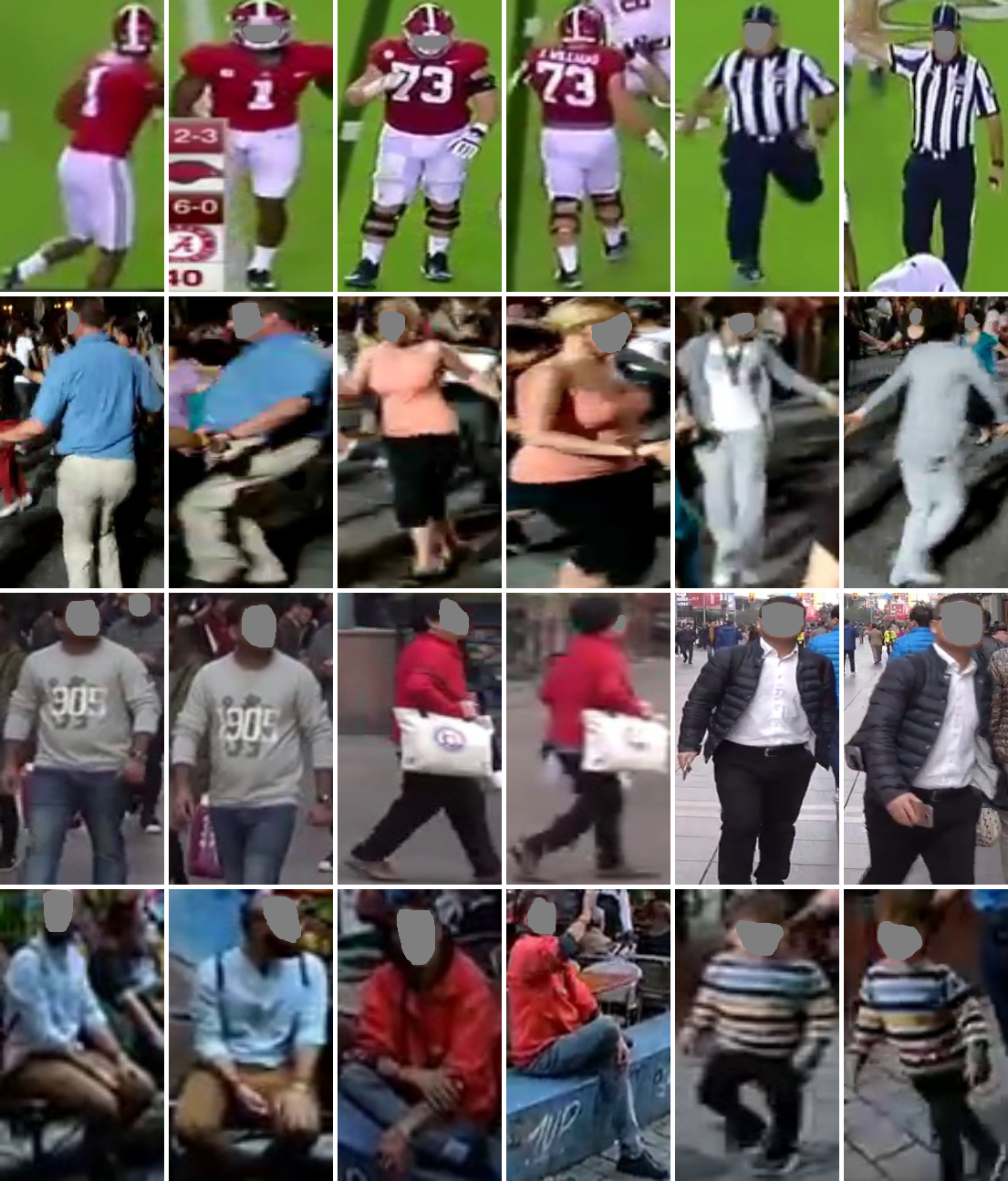}
\end{center}
\vspace{-1.5em}
   \caption{Some example images from our LUPerson dataset, which shows a strong diversity in terms of environment, scene, camera view, lighting, human pose, race, and age.}
\label{fig:LUPerson-demo}
\end{figure}

\subsection{Data Collection and Processing}
To build the dataset, we crawled over 70K streetview videos from YouTube by using queries like ``cityname + streeview (or scene)". To cover a large diversity, we chose the names of top 100 big cities in the world \footnote{Innovation City Index 2019: Top 100 Cities \url{https://www.innovation-cities.com/worlds-top-100-cities-for-innovation-2019/18841/}} and collected about 730 ($680\sim760$) raw videos on average for each city. To ensure a high quality, we further filtered out some invalid videos by checking the following cases: 1) duplicated videos with the same name (YouTube key); 2) videos containing less than 100 frames; 3) static videos (we evenly sample 5 frames and treat this video as a static video if these frames are the same); 4) virtual reality (VR) videos. In total, 50,534 videos are remained after filtering.

We follow the process of building existing person Re-ID datasets, which extract each person instance detected in every image. We use YOLO-v5 \footnote{YOLO-v5 \url{https://github.com/ultralytics/yolov5}} trained on MS-COCO to detect persons in every sampled frame. Considering that some instances only show partially visible body, we apply HRNet~\cite{SunXLW19} to detect the body key-points. Such key-points can be categorized into three types: head, upper body, lower body. In our specification, one person image is regarded as valid if it satisfies the following requirements:  1) head and upper body are visible; 2) lower body is partially visible if either hip or knee exists; 3) The height/width ratio should be larger than 1.5 and less than 5; 4) The detection confidence must be larger than 0.72; 5) The bounding box width must be larger than 48 pixels. Besides, we extract person every 100 frames. After applying these rules, we finally get 4,180,243 person images. To make an estimation of the person number, we adopt the following strategy: we search all the frames in a video to find the frame with the max number of persons and take the max number as the number of person of this video, then we sum up the persons of all the videos and get the number $219,848$. Some example images are given in \Fref{fig:LUPerson-demo}, it shows that our LUPerson has very diverse backgrounds and pose variations. 

\subsection{Comparison with Existing Datasets}
Compared with existing datasets~\cite{gray2008viewpoint,loy2013person,li2014deepreid,Zheng2015ScalablePR,karanam2016comprehensive,zheng2017unlabeled,wei2018person}, LUPerson is superior in the following aspects.

1) \textit{The numbers of images and identities.} To the best of our knowledge, LUPerson is the largest person Re-ID dataset and contains over 4M images of 200K identities, which is $30\times$ larger than MSMT17~\cite{wei2018person}.

2) \textit{Diverse environment.} LUPerson is collected from much diverse environments (containing 46,260 scenes) with both static and dynamic camera views of street, campus, supermarket, sport fields, etc., while existing datasets were normally collected from a few fixed environments (\eg, campus, or no more than 15 scenes~\cite{wei2018person}) and limited camera views. Diverse environments covered by LUPerson are essential for learning a generic Re-ID feature that can be generalized to real applications.

3) \textit{Complex devices.} The raw videos of LUPerson are captured by diverse devices, such as variant types of video recorders, vlogs, smart phones, and surveillance cameras, in contrast to existing datasets which often share the same hardware property.

4) \textit{Lighting variance.} The crawled videos span over a large range of time in a day, including morning, noon, and night, thus causing different lighting changes.

5) \textit{Ethnic and pose diversity.} Since the raw videos are collected from 100 cities across the world, LUPerson has ethnic diversity. Due to the variance of camera views, it also has large pose variations.

\section{Unsupervised Pre-training for Person Re-ID}
\label{sec:ucl}

Based on the LUPerson dataset, we attempt to pre-train an unsupervised model for improving the generalization ability of the learned person Re-ID feature representation. Without loss of generality, we leverage the widely used contrastive learning method MoCoV2~\cite{chen2020improved} as a simple baseline. However, we find such a general setting does not work well on the person Re-ID task. It requires an in-depth study and re-design. 

In this section, we will firstly introduce the contrastive learning method in section~\ref{sec:contrastive_learning}, and then present the key changes specifically for the person Re-ID task in section~\ref{sec:key_factor}. Finally, we will show how the learned representation benefits existing person Re-ID methods in section~\ref{sec:transfer_feature}.

\begin{figure}[t]
    \begin{center}
        \includegraphics[width=1.0\linewidth]{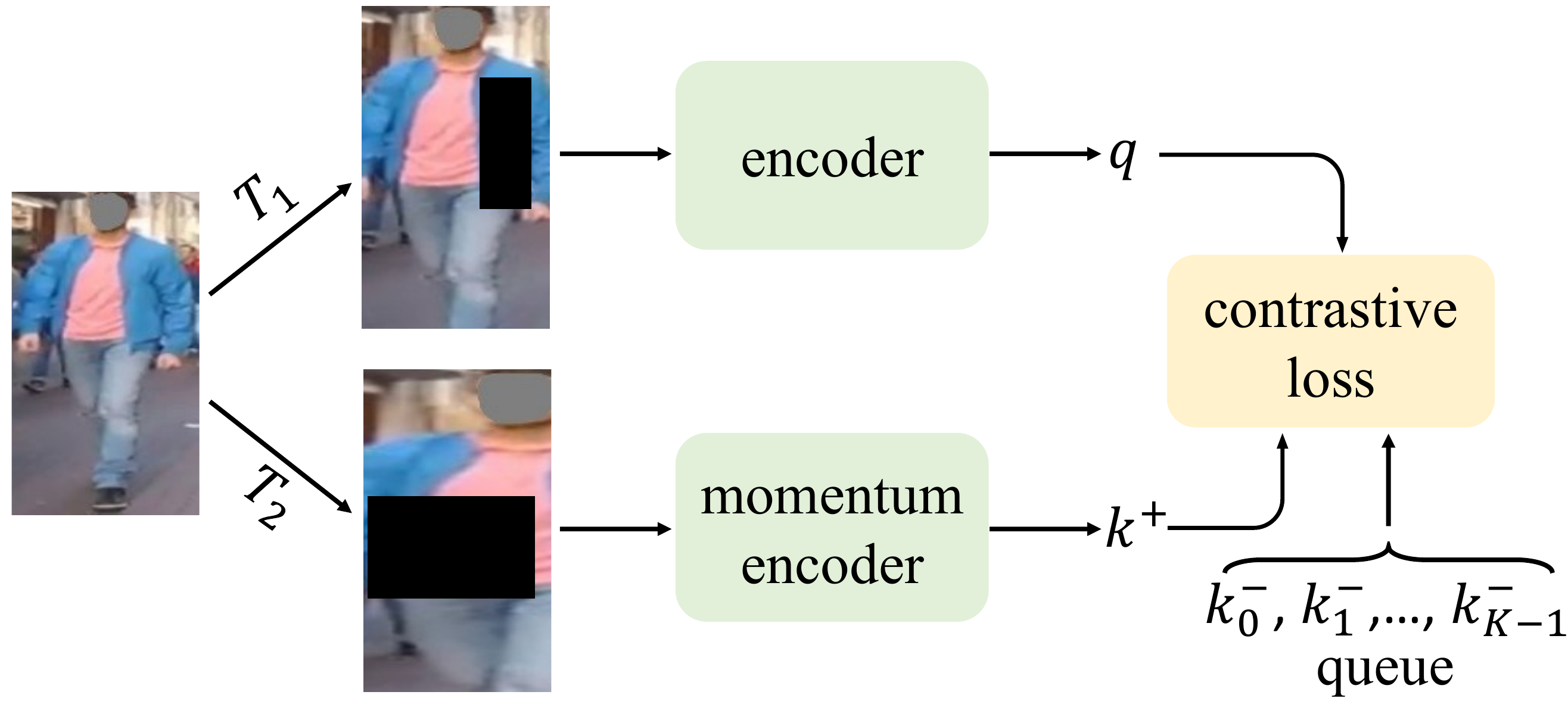}
    \end{center}
  \vspace{-1.0em}
   \caption{Illustration of the Momentum Contrast mechanism for contrastive learning in MoCo~\cite{he2020momentum}.}
\label{fig:moco_v2}
\end{figure}

\subsection{Contrastive Learning}
\label{sec:contrastive_learning}
Contrastive learning can be regarded as a dictionary look-up problem. Given an encoded query $\bm{q}$ and a set of encoded key samples $\{\bm{k}_1, ..., \bm{k}_K\}$ in the dictionary, the contrastive learning is essentially to encourage $\bm{q}$ to be similar to the positive samples $\{\bm{k}_i^+\}$ and dissimilar to the negative samples $\{\bm{k}_i^-\}$. Since no label information is available in unspervised learning, we do not know which are positive or negative. Hence, existing methods (\eg, MoCo~\cite{he2020momentum}) adopt a self-learning strategy. Specifically, as shown in \Fref{fig:moco_v2}, given an image sample $x$, two different augmentations $T_1(x), T_2(x)$ are used to generate two different samples. Then one sample is regarded as query $\bm{q}$ and the other is regarded as the positive sample $\bm{k}^+$. All the samples not augmented from the same image of $\bm{q}$ are then regarded as the negative samples $\bm{k}^-$. Formally, the contrastive loss is defined as:
\begin{equation}
\label{eqn:contrastive}
    \mathcal{L}_{c} = - \text{log}\frac{\text{exp}(\bm{q}\cdot\bm{k}^+ / \tau)}{\text{exp}(\bm{q}\cdot\bm{k}^+/\tau) + \sum_{i=0}^{K-1} \text{exp}(\bm{q}\cdot\bm{k}^-_i/\tau)},
\end{equation}
where $\tau$ is a temperature hyper-parameter, $K$ is the number of negative samples. 

To obtain $(\bm{q},\bm{k})$, separated encoders $f_q$ and $f_k$ are used respectively, and both of them consist of a base feature extractor network (\eg, ResNet50 backbone) and an extra projection head (\eg, two-layer MLP).  Intuitively, the above contrastive loss in Equation~\ref{eqn:contrastive} can be viewed as a $(K+1)$-way classification loss. As demonstrated in existing methods \cite{chen2020simple,wu2018unsupervised}, a large $K$ is beneficial because a rich set of negative samples are covered. One key idea of MoCo is using a queue to maintain a large dictionary and progressively updating the samples in the dictionary with the latest ones. However, updating the key encoder with a large queue by back-propagation is intractable. To solve this issue, another key idea is using momentum updating scheme for $f_k$.
\begin{equation}
    \theta_k := m\theta_k + (1-m) \theta_q,
\end{equation}

where $\theta_k,\theta_q$ are the parameters of $f_k,f_q$, and $m$ is a momentum coefficient.

\subsection{Key Factors to Study}
\label{sec:key_factor}

Compared with general image recognition tasks, person Re-ID is a more challenging task since it needs to seek more fine-grained details for robust person association. Empirically, we observe that directly applying MoCoV2~\cite{chen2020improved} cannot learn a good representation. To improve the performance, we systematically study some key factors: \textit{data augmentation} and \textit{temperature tuning strategy} in the contrastive loss.

\vspace{0.5em}
\noindent\textbf{Data Augmentation.} Data augmentation is crucial to self-supervised contrastive learning. Though the effects of different augmentation operators have been extensively studied for general representation learning \cite{xie2019unsupervised,chen2020simple,he2020momentum}, we find the recommended augmentations are not optimal to the person Re-ID task.

Firstly, we re-investigate the effect of each augmentation operator used in MoCoV2 with respect to the transfer performance on the CUHK03 dataset. Here, we mainly focus on \textit{color distortion augmentations} (including random grayscale, Gaussian blurring and color jitter), since common augmentations like cropping, resizing and flipping are still indispensable to the Re-ID task. The results are shown in \Tref{table:augmentation}. Compared with the ``default" strategy in MoCoV2, we exclude one augmentation at each time. For random grayscale and Gaussian blurring, removing each has limited impact to the performance. Thus, we will not study them in our augmentation. For color jitter, the performance is boosted by $0.6\%$ in terms of \emph{mAP} and $0.6\%$ in terms of \emph{cmc1} after it is disabled. This is because person Re-ID heavily relies on color information (\eg color of clothes, color of bags) for association, while color jitter will harm its performance. 

Then, we revisit the task-specific augmentation \textit{RandomErasing}~\cite{zhong2020random}, which randomly selects a rectangle region in an image and erases its pixels with random values during training. It has been widely used in person Re-ID methods. As we can see in \Tref{table:augmentation}, adding  RandomErasing can achieve about $0.8\%$ gain in terms of \emph{mAP}. Besides, relatively large strength of RandomErasing can help achieve a better result, as shown in \Tref{tab:randomerasing}. Note that the optimal RandomErasing strength on existing person Re-ID method is 0.4, which is smaller than the optimal value (0.6) in our unsupervised pre-training. This indicates that unsupervised pre-training benefits more from stronger data augmentation than vanilla person Re-ID training.

As a summary, we make two important changes in data augmentation for Re-ID contrastive learning: adding RandomErasing with high strength and removing color jitter. 

\vspace{0.5em}
\noindent\textbf{Temperature Tuning Strategy.}
In the contrastive loss, the temperature hyper-parameter $\tau$ effectively weights different examples as shown in Equation~\ref{eqn:contrastive}. A too large value of $\tau$ will reduce the discriminativity between positive samples $\bm{k}^+$ and negative samples $\bm{k}^-$ since it makes $\text{exp}(\bm{q}\cdot\bm{k}^+/\tau)$ close to $\text{exp}(\bm{q}\cdot\bm{k}^-/\tau)$. By contrast, a too small value of $\tau$ would be harmful for the model to learn from hard negative samples, since it makes softmax output more spike towards the positive pair and cannot help the model learn from hard negatives. Hence, an appropriate temperature is critical to help the model learn a discriminative representation.

For this purpose, we search for an optimal $\tau$ with respect to the transfer performance on existing person Re-ID datasets (\eg, CUHK03~\cite{li2014deepreid}). The result is shown in \Tref{tab:temperature}. Using the default $\tau = 0.2$ leads to $\emph{mAP}=71.1$; switching to a smaller $\tau$, the accuracy increases to $\emph{mAP} = 74.7$ $(\tau = 0.07)$. The phenomena consistently appears in other target Re-ID datasets. More interestingly, in general image recognition~\cite{chen2020improved}, the accuracy is reduced from $66.2$ to $62.9$ when $\tau$ varies from $0.2$ to $0.07$. One possible reason is that person Re-ID data is inherently more fine-grained compared with general image recognition data, like ImageNet. In other words, Re-ID data has smaller inter-class variations, which make positive samples close to negative samples. To avoid reducing the discriminitivity, a small temperature is better. It may suggest that a smaller temperature value is more appropriate for contrast learning on fine-grained recognition problems.

\begin{table}[t]
\setlength{\tabcolsep}{2.3mm}
\centering
\small
\begin{tabular}{c|cccccc}
    \shline
    Setting & Default & +RE & -GS & -GB & -CJ & -CJ+RE\\
    \hline
    $mAP$  & 73.4 & 74.2 & 73.2 & 73.3 & 74.0 & 74.7 \\
    \hline
    $cmc1$ & 74.0 & 74.8 & 73.9 & 74.1 & 74.6 & 75.4 \\
	 \shline
\end{tabular}
\caption{Transfer performance on the CUHK03 dataset with different data augmentations. ``+, -"mean with and without, and ``RE, GS, GB, CJ" mean RandomErasing, GrayScale, GaussianBlurring, and ColorJitter respectively.}
\label{table:augmentation}
\end{table}

\begin{table}[!t]
\setlength{\tabcolsep}{3.6mm}
    \centering
    \small
    \begin{tabular}{l|ccccc} 
    \shline
     Max area & 0.0 & 0.2 & 0.4  & 0.6  & 0.8 \\ 
     \hline
     $mAP$  & 73.2 & 74.1 & 74.4 & 74.7 & 73.3 \\
     \hline
     $cmc1$ & 73.8 & 74.3 & 75.3 & 75.4 & 73.7 \\
     \shline
    \end{tabular}
\caption{Transfer performance on the CUHK03 dataset with different RandomErasing strength, i.e., maximum erasing area.}
\label{tab:randomerasing}
\end{table}

\begin{table}[!t]
\setlength{\tabcolsep}{3mm}
    \centering
    \small
    \begin{tabular}{l|cccccc} 
    \shline
     $\tau$ & 0.03 & 0.05 & 0.07 & 0.1  & 0.2  & 0.3 \\ 
     \hline
     $mAP$  & 72.5 & 73.5 & 74.7 & 74.1 & 71.1 & 67.3 \\
     $cmc1$ & 73.4 & 74.0 & 75.4 & 73.9 & 71.5 & 67.4 \\
     \shline
    \end{tabular}
\caption{Transfer performance on the CUHK03 dataset with different temperature $\tau$ in the contrastive loss.}
\label{tab:temperature}
\end{table}

\subsection{Transferring Features}
\label{sec:transfer_feature}
The unsupervised pre-training performed on this large-scale dataset effectively learns a generic person Re-ID feature representation that can benefit existing supervised and unsupervised person Re-ID methods. As mentioned in~\cite{he2020momentum}, features produced by unsupervised pre-training can have different distributions compared with ImageNet supervised pre-training. Besides, the training hyper-parameters (\eg, learning rate) of existing Re-ID methods are also tuned for supervised pre-training models, and thus may not be optimal for unsupervised pre-training models.

To address this issue, we add an extra batch normalization (BN) layer for re-calibration after anywhere the pre-trained feature is used. For example, in MGN~\cite{wang2018learning}, as three heads are appended after the backbone features, we add three extra BN layers before each head. With this strategy, we can use the same hyper-parameters as the ImageNet supervised counterpart. 

\begin{table*}[htb]
\setlength{\tabcolsep}{3.3mm}
    \begin{subtable}[h]{0.5\textwidth}
        \centering
        \begin{tabular}{l|ccc}
        \shline
        pre-train & Trip~\cite{hermans2017defense} & IDE~\cite{zheng2017person} & MGN~\cite{wang2018learning} \\
        \hline
        IN sup.    & 45.2/63.8 & 50.6/55.9 & 70.5/71.2 \\ \hline
        IN unsup.  & 55.5/61.2 & 52.5/57.7 & 67.1/67.0 \\ \hline
        LUP unsup. & \textbf{62.6/67.6} & \textbf{57.6/62.3} & \textbf{74.7/75.4} \\ \shline
        \end{tabular}
        \vspace{-0.12cm}
        \caption{CUHK03}
        \label{tab:improve-cuhk}
    \end{subtable}
    \hfill
    \begin{subtable}[h]{0.5\textwidth}
    \centering
        \begin{tabular}{l|ccc}
        \shline
        pre-train & Trip~\cite{hermans2017defense} & IDE~\cite{zheng2017person} & MGN~\cite{wang2018learning} \\
        \hline
        IN sup.    & 76.2/89.7 & 74.1/90.2 & 87.5/95.1 \\ \hline
        IN unsup.  & 75.1/88.5 & 74.5/89.3 & 88.2/95.3 \\ \hline
        LUP unsup. & \textbf{79.8/71.5} & \textbf{77.9/91.0} & \textbf{91.0/96.4} \\ \shline
        \end{tabular}
        \vspace{-0.12cm}
        \caption{Market1501}
        \label{tab:improve-market}
    \end{subtable}
    \hfill
    \begin{subtable}[h]{0.5\textwidth}
    \centering
        \begin{tabular}{l|ccc}
        \shline
        pre-train & Trip~\cite{hermans2017defense} & IDE~\cite{zheng2017person} & MGN~\cite{wang2018learning} \\
        \hline
        IN sup.    & 65.2/80.7 & 62.8/80.8 & 79.4/89.0 \\ \hline
        IN unsup.  & 65.4/81.1 & 63.4/81.6 & 79.5/89.1 \\ \hline
        LUP unsup. & \textbf{69.8/83.1} & \textbf{65.9/82.2} & \textbf{82.1/91.0} \\ \shline
        \end{tabular}
        \vspace{-0.12cm}
        \caption{DukeMTMC}
        \label{tab:improve-duke}
    \end{subtable}
    \hfill
    \begin{subtable}[h]{0.5\textwidth}
    \centering
        \begin{tabular}{l|ccc}
        \shline
        pre-train & Trip~\cite{hermans2017defense} & IDE~\cite{zheng2017person} & MGN~\cite{wang2018learning} \\
        \hline
        IN sup.    & 34.3/54.8 & 36.2/66.2 & 63.7/85.1 \\ \hline
        IN unsup.  & 34.4/55.4 & 37.6/67.3 & 62.7/84.3 \\ \hline
        LUP unsup. & \textbf{36.6/57.1} & \textbf{39.8/68.9} & \textbf{65.7/85.5} \\ \shline
        \end{tabular}
        \vspace{-0.12cm}
        \caption{MSMT17}
        \label{tab:improve-msmt}
    \end{subtable}
    \vspace{-2mm}
    \caption{Improvement by using different pre-trained models on three representative supervised Re-ID baselines. ``IN sup.", ``IN unsup." refer to supervised and unsupervised pre-trained model on ImageNet, ``LUP unsup." refers to unsupervised pre-trained model on LUPerson. The first number is \emph{mAP} and the second is \emph{cmc1}.}
    \label{tab:impro-sup}
\end{table*}

\begin{table*}[h]
\setlength{\tabcolsep}{1.8mm}
\small
    \begin{subtable}[h]{1.0\textwidth}
        \centering
        \begin{tabular}{l|ccccc|ccccc}
        \shline
        \multirow{2}{*}{pre-train} & \multicolumn{5}{c|}{small-scale} & \multicolumn{5}{c}{few-shot} \\ \cline{2-11} & $10\%$ & $30\%$ & $50\%$ & $70\%$ & $90\%$ & $10\%$ & $30\%$ & $50\%$ & $70\%$ & $90\%$ \\ \hline
        IN sup.    & 53.1/76.9 & 75.2/90.8 & 81.5/93.5 & 84.8/94.5 & 86.9/95.2 & 21.1/41.8 & 68.1/87.6 & 80.2/92.8 & 84.2/94.0 & 86.7/94.6 \\ \hline
        IN unsup.  & 58.4/81.7 & 76.6/91.9 & 82.0/94.1 & 85.4/94.5 & 87.4/95.5 & 18.6/36.1 & 69.3/87.8 & 78.3/90.9 & 84.4/94.1 & 87.1/95.2 \\ \hline
        LUP unsup. & \textbf{64.6/85.5} & \textbf{81.9/93.7} & \textbf{85.8/94.9} & \textbf{88.8/95.9} & \textbf{90.5/96.4} & \textbf{26.4/47.5} & \textbf{78.3/92.1} & \textbf{84.2/93.9} & \textbf{88.4/95.5} & \textbf{90.4/96.3} \\ \shline
    \end{tabular}
    \vspace{-0.12cm}
    \caption{Market1501}
    \label{tab:sd-fw-market}
    \end{subtable}
    
   \begin{subtable}[h]{1.0\textwidth}
        \centering
        \begin{tabular}{l|ccccc|ccccc}
        \shline
        \multirow{2}{*}{pre-train} & \multicolumn{5}{c|}{small-scale} & \multicolumn{5}{c}{few-shot} \\ \cline{2-11} & $10\%$ & $30\%$ & $50\%$ & $70\%$ & $90\%$ & $10\%$ & $30\%$ & $50\%$ & $70\%$ & $90\%$ \\ \hline
        IN sup.    & 45.1/65.3 & 64.7/80.2 & 71.8/84.6 & 75.5/86.8 & 78.0/88.3 & 31.5/47.1 & 65.4/79.8 & 73.9/85.7 & 77.2/87.8 & 79.1/88.8 \\ \hline
        IN unsup.  & 48.1/66.9 & 65.8/80.2 & 72.5/84.4 & 76.3/86.9 & 78.5/88.7 & 32.4/48.0 & 65.3/80.2 & 73.7/85.1 & 77.7/87.8 & 79.4/89.0 \\ \hline
        LUP unsup. & \textbf{53.5/72.0} & \textbf{69.4/81.9} & \textbf{75.6/86.7} & \textbf{78.9/88.2} & \textbf{81.1/90.0} & \textbf{35.8/50.2} & \textbf{72.3/83.8} & \textbf{77.7/87.4} & \textbf{80.8/89.2} & \textbf{82.0/90.6} \\ \shline
    \end{tabular}
    \vspace{-0.12cm}
    \caption{DukeMTMC}
    \label{tab:sd-fw-duke}
    \end{subtable}
    
    \begin{subtable}[h]{1.0\textwidth}
        \centering
        \begin{tabular}{l|ccccc|ccccc}
        \shline
        \multirow{2}{*}{pre-train} & \multicolumn{5}{c|}{small-scale} & \multicolumn{5}{c}{few-shot} \\ \cline{2-11} & $10\%$ & $30\%$ & $50\%$ & $70\%$ & $90\%$ & $10\%$ & $30\%$ & $50\%$ & $70\%$ & $90\%$ \\ \hline
        IN sup.    & 23.2/50.2 & 41.9/70.8 & 50.3/76.9 & 56.9/81.2 & 61.9/84.2 & 14.7/34.1 & 44.5/71.1 & 56.2/79.5 & 60.9/82.8 & 63.4/84.5 \\ \hline
        IN unsup.  & 22.6/48.8 & 40.4/68.7 & 49.0/75.0 & 55.7/79.9 & 60.9/83.0 & 13.2/29.2 & 41.4/67.1 & 53.3/77.6 & 59.1/81.5 & 62.4/83.8 \\ \hline
        LUP unsup. & \textbf{25.5/51.1} & \textbf{44.6/71.4} & \textbf{53.0/77.7} & \textbf{59.5/81.8} & \textbf{63.7/85.0} & \textbf{17.0/36.0} & \textbf{49.0/73.6} & \textbf{57.4/80.5} & \textbf{62.9/83.5} & \textbf{65.0/85.1} \\ \shline
    \end{tabular}
    \vspace{-0.12cm}
    \caption{MSMT17}
    \label{tab:sd-fw-msmt}
    \end{subtable}
    \vspace{-2mm}
    \caption{Performance for small-scale and few-shot setting with MGN method for Market1501, DukeMTMC and MSMT17 respectively.}
    \label{tab:sd-fw}
\end{table*}

\section{Experiments}
\label{sec:exp}

\subsection{Implementation}
\noindent\textbf{Training details.} We train MoCoV2 with Pytorch, and the images are resized to $256\times128$. For image normalization, we use $[0.3525, 0.3106, 0.3140],[0.2660, 0.2522, 0.2505]$ as mean and std, which are calculated from our LUPerson dataset. The pre-training models are trained with $8\times V100$ GPUs for 200 epochs. The initial learning rate is  $0.3$ with batch size 2560. If not specified, the backbone network we used is ResNet50.

\noindent\textbf{Dataset.} To demonstrate the superiority of our pre-training models, we conduct extensive experiments on four target person Re-ID datasets, including CUHK03, Market, DukeMTMC and MSMT17. For CUHK03, following the new protocols proposed in~\cite{zhong2017re}, we split this dataset into two parts: 7,365 images with 767 identities for training and 6,732 images with 700 identities for testing, and we only use its labeled sub-set. For the other three datasets, we use their official settings. 

\noindent\textbf{Evaluation Protocol.} In all the experiments, we follow the standard evaluation metrics: mean Average Precision (mAP) and the Cumulated Matching Characteristics top-1 (cmc1) metric.

\subsection{Improving Supervised Re-ID Methods}
In this section, we show the performance improvement by replacing the supervised pre-trained model on ImageNet with our unsupervised pre-trained model on LUPerson in three representative supervised Re-ID baselines: Trip~\cite{hermans2017defense}, IDE~\cite{zheng2017person} and MGN~\cite{wang2018learning}. The Trip and IDE are our re-implementation based on open source and have comparable or better performance compared with the original papers' claim. For MGN, we use its popular implementation in fast-reid~\cite{he2020fastreid} \footnote{fast-reid: \url{https://github.com/JDAI-CV/fast-reid}}, which can obtain considerable higher performance than what is reported in MGN~\cite{wang2018learning}.

Table \ref{tab:impro-sup} shows the detailed improvements over 4 popular person Re-ID datasets. It can be seen that, equipped with our new per-trained model, all the three methods can have more than $4.2\%, 3.5\%, 2.7\%, 2\%$ improvement in terms of \emph{mAP} on CUHK03, Market1501, DukeMTMC and MSMT17 respectively. As reference, we also compare to the baseline ``unsupervised pre-trained model on ImageNet''. In most cases, the unsupervised ImageNet model is slightly better than the supervised one but much less than our model. On the one hand, it demonstrates the potential of unsupervised pre-training. On the other hand, pre-training on ImageNet is not compatible with pre-training on LUPerson, which further emphasizes the importance of building person-related dataset for person Re-ID.

\subsection{Comparison on Small-scale and Few-shot}

We further study how our pre-trained model benefits the cases where the target dataset has a smaller scale or just a few labels. This is especially important for real applications in which collecting a large labelled Re-ID dataset is so difficult. In this paper, we simulate two typical small dataset settings on the three target datasets: DukeMTMC, Market1501 and MSMT17. CUHK03 is not involved because it is too small.

\begin{itemize}
    \item \textbf{Small-scale.} We randomly select a certain percentage of IDs and all the images belonging to the sampled IDs would be included.
    \item \textbf{Few-shot.} We keep all the identities and randomly sample a certain percentage of images for each ID. During sampling, we try to ensure that each ID has a similar number of images.
\end{itemize}

We use MGN as the baseline method and show the results in Table \ref{tab:sd-fw} by varying the percentage from $10\%\sim 100\%$. As we can see, our pre-training model significantly boosts the performance of MGN in most cases when the training set is small, no matter the identities is less or each identity has few images. 

Specifically, for the ``small-scale" setting on Market1501, which contains only 1,170 images for 75 persons (percentage$=10\%$, more details in supplementary materials), MGN with our pre-training model achieves $64.6$ \emph{mAP} on the testing set, which is $11.5$ \emph{mAP} higher than the ImageNet supervised counterpart. 

For the ``few-shot" setting, our per-training model also boosts the performance of MGN with a remarkable margin, but the performance gain becomes a bit saturated with the decrease of image amounts. One possible reason is that quite limited images for each identity may weaken the ability of hard triplet loss. 

\subsection{Pre-training Data Scale}
\begin{table}[h]
\setlength{\tabcolsep}{1.8mm}
\small
    \centering
    \begin{tabular}{l|cccc}
    \shline
    scale & CUHK03 & Market1501 & DukeMTMC & MSMT17 \\
    \hline
    $12.5\%$ & 69.2/69.0 & 89.5/95.9 & 80.2/89.1 & 61.5/82.3 \\
    $\;\;\;25\%$ & 72.1/71.5 & 90.1/96.2 & 80.9/90.1 & 63.1/83.3 \\
    $\;\;\;50\%$ & 74.1/74.5 & 90.8/96.4 & 81.6/90.4 & 65.5/84.7 \\
    $\;100\%$ & \textbf{74.7/75.4} & \textbf{91.0/96.4} & \textbf{82.1/91.0} & \textbf{65.7/85.5} \\
    \shline
\end{tabular}
\vspace{-1mm}
\caption{Comparison for different pre-training data scale, and the baseline method is MGN.}
\label{tab:data-scale}
\end{table}
In generic image recognition tasks, more powerful models often rely on more high-quality data. Here, we study the impact of pre-training data scale. Specifically, we involve various percentages ($12.5\%$,$25\%$,$50\%$, $100\%$) of LUPerson into unsupervised pre-training and then evaluate the finetuing performance on the target datasets. As shown in Table \ref{tab:data-scale}, the learned representation is much stronger with the increase of the pre-training data scale. But the performance tends to saturate, when the data scale is large enough. A larger network capacity would be necessary to leverage more data for further improving the performance.

\subsection{Improving Unsupervised Re-ID Methods}
\vspace{-2mm}
\begin{table}[h]
\small
    \centering
    \begin{tabular}{l|cc|cc}
    \shline
    \multirow{2}{*}{pre-train} & \multicolumn{2}{c|}{USL} & \multicolumn{2}{c}{UDA} \\
    \cline{2-5} & M & D & D $\rightarrow$ M & M $\rightarrow$ D \\ 
    \hline
    IN sup.    & 72.4/87.8 & 64.9/80.3 & 76.4/90.1 & 67.9/82.3 \\
    IN unsup.  & 72.9/88.6 & 62.6/78.8 & 77.1/90.6 & 66.3/81.6 \\
    LUP unsup. & \textbf{76.2/90.2} & \textbf{67.1/81.6} & \textbf{79.2/91.7} & \textbf{69.1/83.2} \\
    \shline
\end{tabular}
\vspace{-1mm}
\caption{Improvement for unsupervised Re-ID method SpCL. M and D refer to Market1501 and DukeMTMC. Note that, we use the official released code and the performance obtained is slightly lower than the original paper.}
\label{tab:impor-unsup}
\end{table}

Our pre-training model not only benefits the supervised person Re-ID methods, but also is indispensable in existing unsupervised person Re-ID methods. To verify it, we use state-of-the-art unsupervised baseline methods SpCL~\cite{ge2020selfpaced} with its two settings: unsupervised learning (USL) and unsupervised domain adaptation (UDA). We follow the common setting as SpCL~\cite{ge2020selfpaced}, MMT~\cite{ge2019mutual} and MMCL~\cite{wang2020unsupervised}, etc., and evaluate on Market1501 and DukeMTMC. The results are shown in \ref{tab:impor-unsup}. As we can see, our pre-training model can improve $3.8$ \emph{mAP} and $2.2$ \emph{mAP} on Market1501 and DukeMTMC respectively, which is a new state-of-the-art for USL in person Re-ID. For the UDA setting, our model can also boost M $\rightarrow$ D and D $\rightarrow$ M by $2.8\%$ and $1.2\%$ on \emph{mAP}. It further demonstrates the superiority and generality of our pre-training model.

\vspace{-1mm}
\subsection{Comparison with State-of-the-Arts}
\begin{table}
\footnotesize
\setlength{\tabcolsep}{1.1mm}
    \centering
    \begin{tabular}{l|cccc}
    \shline
    Method & CUHK03 & Market1501 & DukeMTMC & MSMT17 \\ 
    \hline
    PCB~\cite{sun2018PCB} (2018) & 57.5/63.7 & 81.6/93.8 & 69.2/83.3 & - \\
    MGN~\cite{wang2018learning} (2018) & 67.4/68.0 & 86.9/95.7 & 78.4/88.7 & - \\
    MGN* & 70.5/71.2 & 87.5/95.1 & 79.4/89.0 & \underline{63.7}/\underline{85.1} \\
    BOT~\cite{luo2019strong} (2019) & - & 85.9/94.5 & 76.4/86.4 & - \\
    DGNet~\cite{zheng2019joint} (2019) & - & 86.0/94.8 & 74.8/86.6 & 52.3/77.2 \\
    IANet~\cite{hou2019interaction} (2019) & - & 83.1/94.4 & 73.4/87.1 & 46.8/75.5 \\
    DSA~\cite{zhang2019densely} (2019) & 75.2/78.9 & 87.6/95.7 & 74.3/86.2 & - \\
    Auto~\cite{quan2019auto} (2019) & 73.0/77.9 & 85.1/94.5 & - & 52.5/78.2 \\
    ABDNet~\cite{chen2019abd} (2019) & - & 88.3/95.6 & 78.6/89.0 & 60.8/82.3 \\
    OSNet~\cite{zhou2019omni} (2019) & 67.8/72.3 & 84.9/94.8 & 73.5/88.6 & 52.9/78.7 \\
    SCAL~\cite{chen2019self} (2019) & 72.3/74.8 & \underline{89.3}/95.8 & 79.6/89.0 & - \\
    P2Net~\cite{guo2019beyond} (2019) & 73.6/78.3 & 85.6/95.2 & 73.1/86.5 & - \\
    MHN~\cite{chen2019mixed} (2019) & 72.4/77.2 & 85.0/95.1 & 77.2/89.1 & - \\
    BDB~\cite{dai2019batch} (2019) & 76.7/79.4 & 86.7/95.3 & 76.0/89.0 & - \\
    SONA~\cite{xia2019second} (2019) & \underline{79.2}/\underline{81.8} & 88.8/95.6 & 78.3/89.4 & - \\
    GCP~\cite{park2020relation} (2020) & 75.6/77.9 & 88.9/95.2 & 78.6/87.9 & - \\
    SAN~\cite{jin2020semantics} (2020) & 76.4/80.1 & 88.0/\underline{96.1} & 75.5/87.9 & 55.7/79.2 \\
    ISP~\cite{zhu2020identity} (2020) & 74.1/76.5 & 88.6/95.3 & \underline{80.0}/\underline{89.6} & - \\
    GASM~\cite{he2020guided} (2020) & - & 84.7/95.3 & 74.4/88.3 & 52.5/79.5 \\
    \hline
     Ours(R50)+BDB & \textbf{79.6/81.9} & 88.1/95.3 & 77.4/88.7 & 52.5/79.1  \\
    Ours(R50)+MGN & 74.7/75.4 & \textbf{91.0}/\textbf{96.4} & \textbf{82.1}/\textbf{91.0} & \textbf{65.7}/\textbf{85.5} \\
    \hline
    MGN(R101) & 73.5/74.6 & 89.0/95.8 & 80.9/89.8 & 66.0/85.7 \\
    Ours(R101)+MGN & 76.9/77.6 & \textbf{92.0/97.0} & \textbf{84.1/91.9} & \textbf{68.8/86.6} \\
    \shline
\end{tabular}
\vspace{-1mm}
\caption{Comparison with state of the arts. MGN* refers to the re-implementation of MGN in fast-reid. The best is marked as bold and the second is underlined.}
\label{tab:comp-sota}
\end{table}

We compare our results with existing state-of-the-art Re-ID methods in Table \ref{tab:comp-sota}. Note that we do not apply any post-processing method like Re-Rank~\cite{zhong2017re} in our approach. As we can see, we achieve state-of-the-art performance on Market1501, DukeMTMC and MSMT17 with considerable advantages, by simply applying our pre-training ResNet50 model on MGN. On CUHK03, there is a noticeable gap between the baseline MGN and the state-of-the-art SONA~\cite{xia2019second}. Thus, we also apply our pre-trained model upon BDB \cite{dai2019batch}, which achieves $2.9$ \emph{mAP} gains over BDB and helps beat SONA~\cite{xia2019second}. 

We also test our method on a large backbone ResNet101. Compared to results of ResNet50, we can see that a stronger backbone can learn better representation features, yielding better performance.

\vspace{-0.25cm}

\section{Conclusion}
\label{sec:concl}
This paper releases a new large-scale person Re-ID dataset ``LUPerson", which may address the issue of limited scale and diversities in all existing Re-ID datasets. Based on such an unlabelled dataset, we make the first attempt to utilize  unsupervised pre-training to learn a general person Re-ID feature representation. Experiments demonstrated the effectiveness and generalization ability of our pre-training model on both supervised and unsupervised Re-ID approaches. Furthermore, it helps achieve bigger gains in small-scale or few-shot Re-ID datasets. Certainly, there are some interesting topics motivated in this direction, including leveraging the temporal information of videos into the pre-training and developing an end-to-end unsupervised Re-ID model which can beat supervised ones.

\vspace{2mm}
\noindent \textbf{Acknowledgement.} This work is partially supported by the National Natural Science Foundation of China (NSFC, 61836011).

{\small
\bibliographystyle{ieee_fullname}
\bibliography{egbib}

\begin{thebibliography}{10}\itemsep=-1pt

\bibitem{caron2020unsupervised}
Mathilde Caron, Ishan Misra, Julien Mairal, Priya Goyal, Piotr Bojanowski, and
  Armand Joulin.
\newblock Unsupervised learning of visual features by contrasting cluster
  assignments.
\newblock In {\em Thirty-fourth Conference on Neural Information Processing
  Systems (NeurIPS)}, 2020.

\bibitem{chen2019mixed}
Binghui Chen, Weihong Deng, and Jiani Hu.
\newblock Mixed high-order attention network for person re-identification.
\newblock In {\em Proceedings of the IEEE International Conference on Computer
  Vision}, pages 371--381, 2019.

\bibitem{chen2018group}
Dapeng Chen, Dan Xu, Hongsheng Li, Nicu Sebe, and Xiaogang Wang.
\newblock Group consistent similarity learning via deep crf for person
  re-identification.
\newblock In {\em Proceedings of the IEEE Conference on Computer Vision and
  Pattern Recognition}, pages 8649--8658, 2018.

\bibitem{chen2019self}
Guangyi Chen, Chunze Lin, Liangliang Ren, Jiwen Lu, and Jie Zhou.
\newblock Self-critical attention learning for person re-identification.
\newblock In {\em Proceedings of the IEEE International Conference on Computer
  Vision}, pages 9637--9646, 2019.

\bibitem{chen2019abd}
Tianlong Chen, Shaojin Ding, Jingyi Xie, Ye Yuan, Wuyang Chen, Yang Yang, Zhou
  Ren, and Zhangyang Wang.
\newblock Abd-net: Attentive but diverse person re-identification.
\newblock In {\em Proceedings of the IEEE International Conference on Computer
  Vision}, pages 8351--8361, 2019.

\bibitem{chen2020simple}
Ting Chen, Simon Kornblith, Mohammad Norouzi, and Geoffrey Hinton.
\newblock A simple framework for contrastive learning of visual
  representations.
\newblock {\em arXiv preprint arXiv:2002.05709}, 2020.

\bibitem{chen2020big}
Ting Chen, Simon Kornblith, Kevin Swersky, Mohammad Norouzi, and Geoffrey
  Hinton.
\newblock Big self-supervised models are strong semi-supervised learners.
\newblock {\em arXiv preprint arXiv:2006.10029}, 2020.

\bibitem{chen2020mocov2}
Xinlei Chen, Haoqi Fan, Ross Girshick, and Kaiming He.
\newblock Improved baselines with momentum contrastive learning.
\newblock {\em arXiv preprint arXiv:2003.04297}, 2020.

\bibitem{chen2020improved}
Xinlei Chen, Haoqi Fan, Ross Girshick, and Kaiming He.
\newblock Improved baselines with momentum contrastive learning.
\newblock {\em arXiv preprint arXiv:2003.04297}, 2020.

\bibitem{dai2019batch}
Zuozhuo Dai, Mingqiang Chen, Xiaodong Gu, Siyu Zhu, and Ping Tan.
\newblock Batch dropblock network for person re-identification and beyond.
\newblock In {\em Proceedings of the IEEE International Conference on Computer
  Vision}, pages 3691--3701, 2019.

\bibitem{dollar2014fast}
Piotr Doll{\'a}r, Ron Appel, Serge Belongie, and Pietro Perona.
\newblock Fast feature pyramids for object detection.
\newblock {\em IEEE transactions on pattern analysis and machine intelligence},
  36(8):1532--1545, 2014.

\bibitem{felzenszwalb2009object}
Pedro~F Felzenszwalb, Ross~B Girshick, David McAllester, and Deva Ramanan.
\newblock Object detection with discriminatively trained part-based models.
\newblock {\em IEEE transactions on pattern analysis and machine intelligence},
  32(9):1627--1645, 2009.

\bibitem{fu2020improving}
Dengpan Fu, Bo Xin, Jingdong Wang, Dongdong Chen, Jianmin Bao, Gang Hua, and
  Houqiang Li.
\newblock Improving person re-identification with iterative impression
  aggregation.
\newblock {\em IEEE Transactions on Image Processing}, 29:9559--9571, 2020.

\bibitem{ge2019mutual}
Yixiao Ge, Dapeng Chen, and Hongsheng Li.
\newblock Mutual mean-teaching: Pseudo label refinery for unsupervised domain
  adaptation on person re-identification.
\newblock In {\em International Conference on Learning Representations}, 2019.

\bibitem{ge2020selfpaced}
Yixiao Ge, Feng Zhu, Dapeng Chen, Rui Zhao, and Hongsheng Li.
\newblock Self-paced contrastive learning with hybrid memory for domain
  adaptive object re-id.
\newblock In {\em Advances in Neural Information Processing Systems}, 2020.

\bibitem{gray2008viewpoint}
Douglas Gray and Hai Tao.
\newblock Viewpoint invariant pedestrian recognition with an ensemble of
  localized features.
\newblock In {\em European conference on computer vision}, pages 262--275.
  Springer, 2008.

\bibitem{grill2020bootstrap}
Jean-Bastien Grill, Florian Strub, Florent Altch{\'e}, Corentin Tallec,
  Pierre~H Richemond, Elena Buchatskaya, Carl Doersch, Bernardo~Avila Pires,
  Zhaohan~Daniel Guo, Mohammad~Gheshlaghi Azar, et~al.
\newblock Bootstrap your own latent: A new approach to self-supervised
  learning.
\newblock {\em arXiv preprint arXiv:2006.07733}, 2020.

\bibitem{guo2019beyond}
Jianyuan Guo, Yuhui Yuan, Lang Huang, Chao Zhang, Jin-Ge Yao, and Kai Han.
\newblock Beyond human parts: Dual part-aligned representations for person
  re-identification.
\newblock In {\em Proceedings of the IEEE International Conference on Computer
  Vision}, pages 3642--3651, 2019.

\bibitem{he2020momentum}
Kaiming He, Haoqi Fan, Yuxin Wu, Saining Xie, and Ross Girshick.
\newblock Momentum contrast for unsupervised visual representation learning.
\newblock In {\em Proceedings of the IEEE/CVF Conference on Computer Vision and
  Pattern Recognition}, pages 9729--9738, 2020.

\bibitem{he2020fastreid}
Lingxiao He, Xingyu Liao, Wu Liu, Xinchen Liu, Peng Cheng, and Tao Mei.
\newblock Fastreid: A pytorch toolbox for general instance re-identification.
\newblock {\em arXiv preprint arXiv:2006.02631}, 2020.

\bibitem{he2020guided}
Lingxiao He and Wu Liu.
\newblock Guided saliency feature learning for person re-identification in
  crowded scenes.
\newblock In {\em European Conference on Computer Vision}, pages 357--373.
  Springer, 2020.

\bibitem{hermans2017defense}
Alexander Hermans, Lucas Beyer, and Bastian Leibe.
\newblock In defense of the triplet loss for person re-identification.
\newblock {\em arXiv preprint arXiv:1703.07737}, 2017.

\bibitem{hou2019interaction}
Ruibing Hou, Bingpeng Ma, Hong Chang, Xinqian Gu, Shiguang Shan, and Xilin
  Chen.
\newblock Interaction-and-aggregation network for person re-identification.
\newblock In {\em Proceedings of the IEEE Conference on Computer Vision and
  Pattern Recognition}, pages 9317--9326, 2019.

\bibitem{jin2020semantics}
Xin Jin, Cuiling Lan, Wenjun Zeng, Guoqiang Wei, and Zhibo Chen.
\newblock Semantics-aligned representation learning for person
  re-identification.
\newblock In {\em AAAI}, pages 11173--11180, 2020.

\bibitem{karanam2016comprehensive}
Srikrishna Karanam, Mengran Gou, Ziyan Wu, Angels Rates-Borras, Octavia Camps,
  and Richard~J Radke.
\newblock A comprehensive evaluation and benchmark for person
  re-identification: Features, metrics, and datasets.
\newblock {\em arXiv preprint arXiv:1605.09653}, 2(3):5, 2016.

\bibitem{li2014deepreid}
Wei Li, Rui Zhao, Tong Xiao, and Xiaogang Wang.
\newblock Deepreid: Deep filter pairing neural network for person
  re-identification.
\newblock In {\em Proceedings of the IEEE conference on computer vision and
  pattern recognition}, pages 152--159, 2014.

\bibitem{lin2019bottom}
Yutian Lin, Xuanyi Dong, Liang Zheng, Yan Yan, and Yi Yang.
\newblock A bottom-up clustering approach to unsupervised person
  re-identification.
\newblock In {\em Proceedings of the AAAI Conference on Artificial
  Intelligence}, volume~33, pages 8738--8745, 2019.

\bibitem{loy2013person}
Chen~Change Loy, Chunxiao Liu, and Shaogang Gong.
\newblock Person re-identification by manifold ranking.
\newblock In {\em 2013 IEEE International Conference on Image Processing},
  pages 3567--3571. IEEE, 2013.

\bibitem{luo2019strong}
Hao Luo, Wei Jiang, Youzhi Gu, Fuxu Liu, Xingyu Liao, Shenqi Lai, and Jianyang
  Gu.
\newblock A strong baseline and batch normalization neck for deep person
  re-identification.
\newblock {\em IEEE Transactions on Multimedia}, 2019.

\bibitem{park2020relation}
Hyunjong Park and Bumsub Ham.
\newblock Relation network for person re-identification.
\newblock In {\em Proceedings of the AAAI Conference on Artificial
  Intelligence}, volume~34, pages 11839--11847, 2020.

\bibitem{quan2019auto}
Ruijie Quan, Xuanyi Dong, Yu Wu, Linchao Zhu, and Yi Yang.
\newblock Auto-reid: Searching for a part-aware convnet for person
  re-identification.
\newblock In {\em Proceedings of the IEEE International Conference on Computer
  Vision}, pages 3750--3759, 2019.

\bibitem{ren2015faster}
Shaoqing Ren, Kaiming He, Ross Girshick, and Jian Sun.
\newblock Faster r-cnn: Towards real-time object detection with region proposal
  networks.
\newblock In {\em Advances in neural information processing systems}, pages
  91--99, 2015.

\bibitem{shen2018person}
Yantao Shen, Hongsheng Li, Shuai Yi, Dapeng Chen, and Xiaogang Wang.
\newblock Person re-identification with deep similarity-guided graph neural
  network.
\newblock In {\em Proceedings of the European conference on computer vision
  (ECCV)}, pages 486--504, 2018.

\bibitem{suh2018part}
Yumin Suh, Jingdong Wang, Siyu Tang, Tao Mei, and Kyoung Mu~Lee.
\newblock Part-aligned bilinear representations for person re-identification.
\newblock In {\em Proceedings of the European Conference on Computer Vision
  (ECCV)}, pages 402--419, 2018.

\bibitem{SunXLW19}
Ke Sun, Bin Xiao, Dong Liu, and Jingdong Wang.
\newblock Deep high-resolution representation learning for human pose
  estimation.
\newblock In {\em CVPR}, 2019.

\bibitem{sun2018PCB}
Yifan Sun, Liang Zheng, Yi Yang, Qi Tian, and Shengjin Wang.
\newblock Beyond part models: Person retrieval with refined part pooling (and a
  strong convolutional baseline).
\newblock In {\em ECCV}, 2018.

\bibitem{wang2020unsupervised}
Dongkai Wang and Shiliang Zhang.
\newblock Unsupervised person re-identification via multi-label classification.
\newblock In {\em Proceedings of the IEEE/CVF Conference on Computer Vision and
  Pattern Recognition}, pages 10981--10990, 2020.

\bibitem{wang2018learning}
Guanshuo Wang, Yufeng Yuan, Xiong Chen, Jiwei Li, and Xi Zhou.
\newblock Learning discriminative features with multiple granularities for
  person re-identification.
\newblock In {\em 2018 ACM Multimedia Conference on Multimedia Conference},
  pages 274--282. ACM, 2018.

\bibitem{wei2018person}
Longhui Wei, Shiliang Zhang, Wen Gao, and Qi Tian.
\newblock Person transfer gan to bridge domain gap for person
  re-identification.
\newblock In {\em Proceedings of the IEEE Conference on Computer Vision and
  Pattern Recognition}, pages 79--88, 2018.

\bibitem{wu2018unsupervised}
Zhirong Wu, Yuanjun Xiong, Stella~X Yu, and Dahua Lin.
\newblock Unsupervised feature learning via non-parametric instance
  discrimination.
\newblock In {\em Proceedings of the IEEE Conference on Computer Vision and
  Pattern Recognition}, pages 3733--3742, 2018.

\bibitem{xia2019second}
Bryan~Ning Xia, Yuan Gong, Yizhe Zhang, and Christian Poellabauer.
\newblock Second-order non-local attention networks for person
  re-identification.
\newblock In {\em Proceedings of the IEEE International Conference on Computer
  Vision}, pages 3760--3769, 2019.

\bibitem{xie2019unsupervised}
Qizhe Xie, Zihang Dai, Eduard Hovy, Minh-Thang Luong, and Quoc~V Le.
\newblock Unsupervised data augmentation for consistency training.
\newblock {\em arXiv preprint arXiv:1904.12848}, 2019.

\bibitem{zhang2019densely}
Zhizheng Zhang, Cuiling Lan, Wenjun Zeng, and Zhibo Chen.
\newblock Densely semantically aligned person re-identification.
\newblock In {\em Proceedings of the IEEE Conference on Computer Vision and
  Pattern Recognition}, pages 667--676, 2019.

\bibitem{Zheng2015ScalablePR}
Liang Zheng, Liyue Shen, Lu Tian, Shengjin Wang, Jingdong Wang, and Qi Tian.
\newblock Scalable person re-identification: A benchmark.
\newblock {\em 2015 IEEE International Conference on Computer Vision (ICCV)},
  pages 1116--1124, 2015.

\bibitem{zheng2017person}
Liang Zheng, Hengheng Zhang, Shaoyan Sun, Manmohan Chandraker, Yi Yang, and Qi
  Tian.
\newblock Person re-identification in the wild.
\newblock In {\em Proceedings of the IEEE Conference on Computer Vision and
  Pattern Recognition}, pages 1367--1376, 2017.

\bibitem{zheng2019joint}
Zhedong Zheng, Xiaodong Yang, Zhiding Yu, Liang Zheng, Yi Yang, and Jan Kautz.
\newblock Joint discriminative and generative learning for person
  re-identification.
\newblock In {\em Proceedings of the IEEE conference on computer vision and
  pattern recognition}, pages 2138--2147, 2019.

\bibitem{zheng2017unlabeled}
Zhedong Zheng, Liang Zheng, and Yi Yang.
\newblock Unlabeled samples generated by gan improve the person
  re-identification baseline in vitro.
\newblock In {\em Proceedings of the IEEE International Conference on Computer
  Vision}, 2017.

\bibitem{zhong2017re}
Zhun Zhong, Liang Zheng, Donglin Cao, and Shaozi Li.
\newblock Re-ranking person re-identification with k-reciprocal encoding.
\newblock In {\em Proceedings of the IEEE Conference on Computer Vision and
  Pattern Recognition}, pages 1318--1327, 2017.

\bibitem{zhong2020random}
Zhun Zhong, Liang Zheng, Guoliang Kang, Shaozi Li, and Yi Yang.
\newblock Random erasing data augmentation.
\newblock In {\em Proceedings of the AAAI Conference on Artificial Intelligence
  (AAAI)}, 2020.

\bibitem{zhou2019omni}
Kaiyang Zhou, Yongxin Yang, Andrea Cavallaro, and Tao Xiang.
\newblock Omni-scale feature learning for person re-identification.
\newblock In {\em Proceedings of the IEEE International Conference on Computer
  Vision}, pages 3702--3712, 2019.

\bibitem{zhu2020identity}
Kuan Zhu, Haiyun Guo, Zhiwei Liu, Ming Tang, and Jinqiao Wang.
\newblock Identity-guided human semantic parsing for person re-identification.
\newblock {\em ECCV}, 2020.

\end{thebibliography}
}

\end{document}